\documentclass[11pt]{article}
\usepackage{arxiv}

\usepackage{doi}

\usepackage[T1]{fontenc}
\usepackage{lmodern}
\usepackage{microtype}
\usepackage{geometry}
\usepackage{graphicx}
\usepackage{booktabs}
\usepackage{longtable}
\usepackage{array}
\usepackage{multirow}
\usepackage{tabularx}
\usepackage{amsmath,amssymb}
\usepackage{hyperref}
\usepackage{xcolor}
\usepackage{enumitem}
\usepackage{caption}
\usepackage{float}
\usepackage{url}
\usepackage{subcaption}

\usepackage{ragged2e} 

\newcolumntype{P}[1]{>{\RaggedRight\arraybackslash\hyphenpenalty=10000\exhyphenpenalty=10000\sloppy}p{#1}}

\newcolumntype{M}[1]{>{\RaggedRight\arraybackslash\hyphenpenalty=10000\exhyphenpenalty=10000\sloppy}m{#1}}

\usepackage{fancyhdr}
\fancyheadoffset{0pt}
\hypersetup{
    colorlinks=true,
    linkcolor=blue,
    citecolor=blue,
    urlcolor=blue,
    pdftitle={PR3DICTR: Modular AI framework for medical 3D image-based detection and outcome prediction},
    pdfauthor={D.C. MacRae, L. van der Hoek, R. van der Wal, S.P.M. de Vette, H. Neh, B. Ma, P.M.A. van Ooijen, L.V. van Dijk}
}

\title{\textbf{PR\textcolor{cyan}{3D}ICTR}: A modular AI framework for medical 3D image-based detection and outcome prediction}

\author{
Daniel C. MacRae\thanks{These authors contributed equally to this work},~\ 
Luuk van der Hoek\footnotemark[1],~\ 
Robert van der Wal,~\
Suzanne P.M. de Vette,~ \\
Hendrike Neh,~\ 
Baoqiang Ma,~\
Peter M.A. van Ooijen,~\ 
Lisanne V. van Dijk \\[0.5em]
\small Department of Radiation Oncology, \\
\small University Medical Center Groningen, \\
\small University of Groningen, \\
\small Groningen, the Netherlands
}

\date{}

\begin{document}

\maketitle

\begin{abstract}
Three-dimensional medical image data and computer-aided decision making, particularly using deep learning, are becoming increasingly important in the medical field. To aid in these developments we introduce \textbf{PR3DICTR}: \textbf{P}latform for \textbf{R}esearch in \textbf{3D} \textbf{I}mage \textbf{C}lassification and s\textbf{T}andardised t\textbf{R}aining. Built using community-standard distributions (PyTorch and MONAI), PR3DICTR provides an open-access, flexible and convenient framework for prediction model development, with an explicit focus on classification using three-dimensional medical image data. By combining modular design principles and standardization, it aims to alleviate developmental burden whilst retaining adjustability. It provides users with a wealth of pre-established functionality, for instance in model architecture design options, hyper-parameter solutions and training methodologies, but still gives users the opportunity and freedom to ``plug in'' their own solutions or modules. PR3DICTR can be applied to any binary or event-based three-dimensional classification task and can work with as little as two lines of code.
\end{abstract}

\section{Introduction}

\begin{figure}[H]
    \centering
    \includegraphics[width=0.9\textwidth]{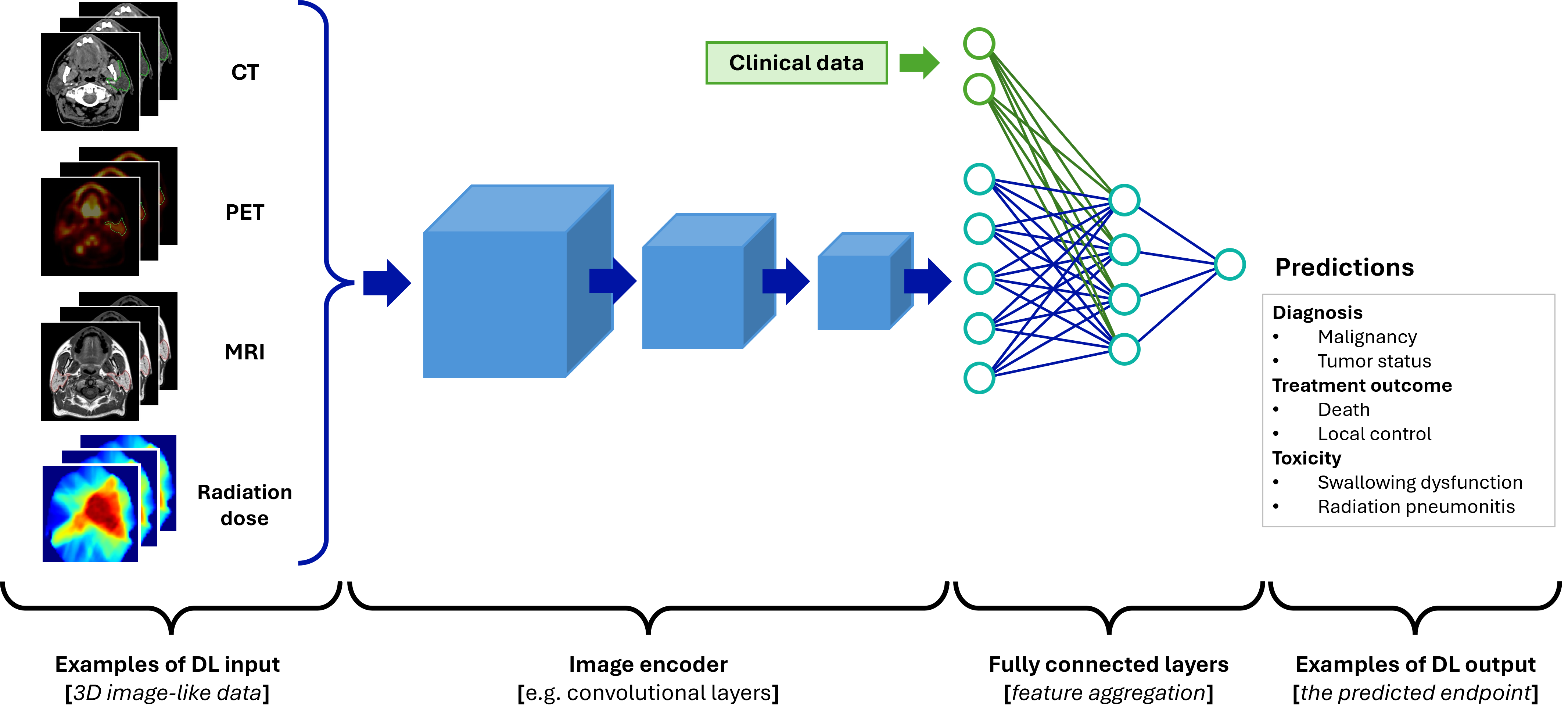}
    \caption{Example of a binary classification deep learning model workflow using 3D medical imaging.}
    \label{fig:workflow_binary}
\end{figure}

In recent years, three-dimensional (3D) multi-modal medical imaging has become increasingly important for diagnosis and treatment of patients \cite{Rong2024}. These scans provide both anatomical and functional information that can guide treatment selection and predict outcomes such as survival, risk of (surgical) complications or treatment failure \cite{Vagvala2022,Saba2025}. While clinicians traditionally rely on guidelines built on discrete clinical variables (e.g., TNM staging) and semantic or manually derived imaging features (e.g., ``hypo-dense lesions'' or aortic diameter measurements), advanced image processing and artificial intelligence (AI) now enable automated extraction of features directly from full 3D scans \cite{Aiello2019,vanGalen2025,Huang2024,Erasmus2023,Wehbe2024}. Deep learning (DL) models, such as convolutional neural networks (CNNs, Figure~\ref{fig:workflow_binary}) and transformers, have shown great potential for capturing intricate patterns in medical imaging data and associating them with clinical outcomes \cite{Huang2024,Wehbe2024}. However, their development often requires substantial computational resources and technical expertise, leading to diverse and often non-standardized workflows both across and even within research groups \cite{Pati2023}. Consequently, there remains a need for simple, standardized, and modular tools that make working with 3D image-based deep learning model development more accessible.

When researchers begin developing machine learning or deep learning models in Python, a wide range of frameworks are available. For maximum flexibility, PyTorch offers a powerful foundation and the tools to develop diverse deep learning applications, from large language models to financial forecasting models \cite{Paszke2019}. Within the medical imaging domain, the Medical Open Network for Artificial Intelligence (MONAI) builds on PyTorch to provide specialized tools for 2D and 3D medical data, while maintaining much of PyTorch's flexibility \cite{Cardoso2022}. In contrast, no-code frameworks such as Ludwig allow users to define models via configuration files rather than custom code, specifying inputs, architectures, and outputs directly \cite{Molino2019}. Overall, existing solutions tend to be either too broad to be readily applied to specific medical imaging classification tasks or too narrowly focused to be adaptable to the requirements of individual outcome prediction models. There remains a need for a balanced framework that combines simplicity, modularity, and standardization for classification modelling with 3D medical data.

We address these needs in the development of a framework that offers high modularity while retaining simplicity and standardization for deep learning (DL) development projects based on 3D medical imaging data: the PR3DICTR framework (\textbf{P}latform for \textbf{R}esearch in \textbf{3D} \textbf{I}mage \textbf{C}lassification and s\textbf{T}andardised tRaining). Through standardizing crucial tasks such as data loading, model training, hyper-parameter optimisation, and model evaluation, we aim to alleviate the development burden and increase consistency for and between scientists.

Additionally, the framework incorporates standard solutions for common medical data challenges, including missing value and imbalance handling, regularization using 3D image augmentation, and multi-modality imaging with optional input-specific preprocessing together with tabular data integration. Built in a modular fashion using standard PyTorch and MONAI components, the framework enables users to replace or extend individual workflow elements (such as custom image feature extractors) while providing an end-to-end solution for 3D medical image model development and evaluation that can be readily adapted to new prediction tasks.

This paper describes the PR3DICTR framework and its application in an open-source dataset.

\section{Methods}

\begin{figure}[H]
    \centering
    \includegraphics[width=0.95\textwidth]{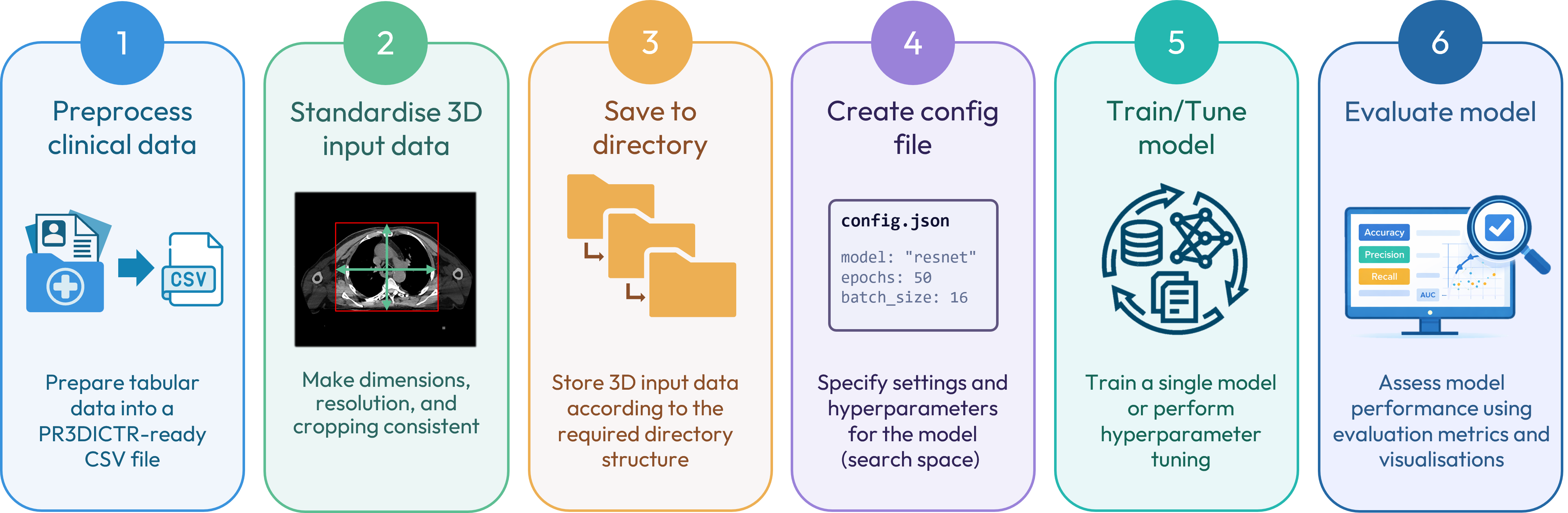}
    \caption{Overall PR3DICTR workflow. An example notebook for steps 1--3 can be found in the repository, and another notebook demonstrates steps 4--6.}
    \label{fig:overall_workflow}
\end{figure}

The overall PR3DICTR workflow is depicted in Figure~\ref{fig:overall_workflow}. The pipeline encompasses 1) data preprocessing, 2) image standardisation, 3) data organisation, 4) configuration setup, 5) model training, and 6) evaluation.  These steps are described in detail in the following subsections. The complete code for PR3DICTR can be found at \url{https://github.com/DLinRadiotherapyUMCG/PR3DICTR}.

\subsection{Prerequisite: Data curation}

Medical imaging data is highly varied, with a multitude of modalities (CT, PET, MRI, radiation dose, etc.), where each individual image modality usually has a different number of slices, intensity ranges and resolution. Deep-learning models typically require a consistent 3D input with fixed dimensions and intensities for effective training and inference.

\begin{figure}[ht]
    \centering
    \includegraphics[width=0.75\textwidth]{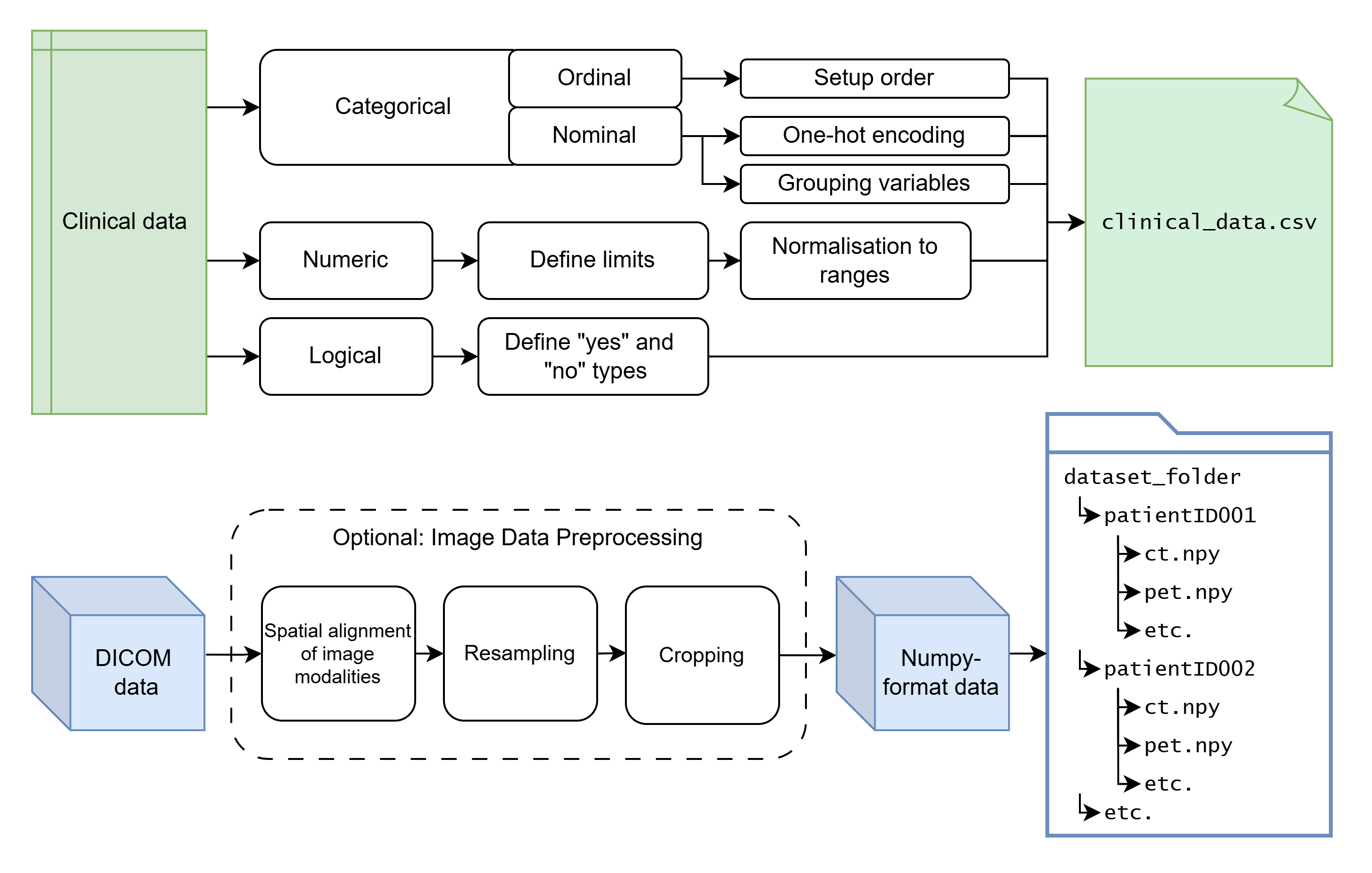}
    \caption{Schematic of a data pre-processing pipeline for preparing datasets to be used in the PR3DICTR framework.}
    \label{fig:data_curation}
\end{figure}

Within this technical note, we distinguish between (1) dataset curation, which converts raw DICOM data into structured NumPy datasets, and (2) input preprocessing, which prepares curated data for model training via cropping, value clipping and normalization. The PR3DICTR framework includes only the latter and assumes that users have already curated their data into a (semi) standardized format, which is outlined in Figure~\ref{fig:data_curation}.

Data curation consists of three steps:
\begin{enumerate}
    \item \textbf{Preprocess clinical/tabular data (Figure~\ref{fig:overall_workflow}, step 1).} The PR3DICTR framework requires a CSV file containing clinical data and endpoints for each patient. Three standardised columns are mandatory: \texttt{PatientID} (linking clinical and imaging data), \texttt{Split} (either \texttt{train\_val} or \texttt{test}, indicating the set), and at least one column containing labels with which to train the model. Missing labels can be indicated using a missing indicator, for which the default is \texttt{-1}. During model training and evaluation, any endpoints set to \texttt{-1} will be ignored (for instance when calculating the loss or evaluation metrics). Furthermore, the CSV file can contain any tabular features to include as a model input (e.g. patient age). \\
    \item \textbf{Standardise 3D input data (Figure~\ref{fig:overall_workflow}, step 2).} The PR3DICTR framework has two requirements for any 3D volumetric data to be used as model input. First, all volumetric data must be of the same dimension (i.e., size). Secondly, volume data is required to be stored as NumPy (\texttt{.npy}) files. It is also recommended in this step to crop the volumetric data to the same field-of-view (for example, so that CTs and MRIs for each patient are spatially aligned). \\
    \item \textbf{Save data into directory structure (Figure~\ref{fig:overall_workflow}, step 3).} Once the volumetric data is prepared, it needs to be stored in a fixed folder structure based on the \texttt{PatientID}s. One main data folder should contain subfolders, one per \texttt{PatientID} in the dataset; these subfolders should then contain the volume data named per volume (for instance \texttt{data/PatientID001/PET.npy}, \texttt{data/PatientID001/CT.npy}, \ldots), as illustrated in Figure~\ref{fig:data_curation}. \\
\end{enumerate}

\subsection{Modularity}

PR3DICTR is designed with a modular nature, allowing each component---data loading, model architecture, K-fold cross-validation and training, and evaluation---to be used independently or in combination for different research experiments. This flexible design enables researchers to swap, extend, or customize individual modules without affecting the rest of the pipeline, supporting rapid prototyping, reproducibility, and exploration of new methodologies. In the following subsections, we describe the main categories of modules in more detail.

\subsection{Framework \& user interaction}

One of the aims of PR3DICTR is to simplify the development process for new 3D deep learning--based solutions to volume-based medical problems. This simplification is achieved primarily using configuration files (``configs''). When starting a new project, users prepare their dataset and a corresponding config file (Figure~\ref{fig:overall_workflow}, step 4). Subsequently, only two lines of code are required: one to load the config and one to execute the experiment. The config is passed through the model training pipeline as a dictionary and contains all parameters required for training and evaluation.

The config file serves as the central hub for defining every aspect of the model definition and training setup, allowing users to modify hyperparameters and other settings with ease. Among other elements (see Table~\ref{tab:options}), it specifies optional model input image preprocessing steps, data augmentations, model architecture and size, training hyperparameters, and evaluation metrics.

\begin{figure}[ht]
    \centering
    \includegraphics[width=0.9\textwidth]{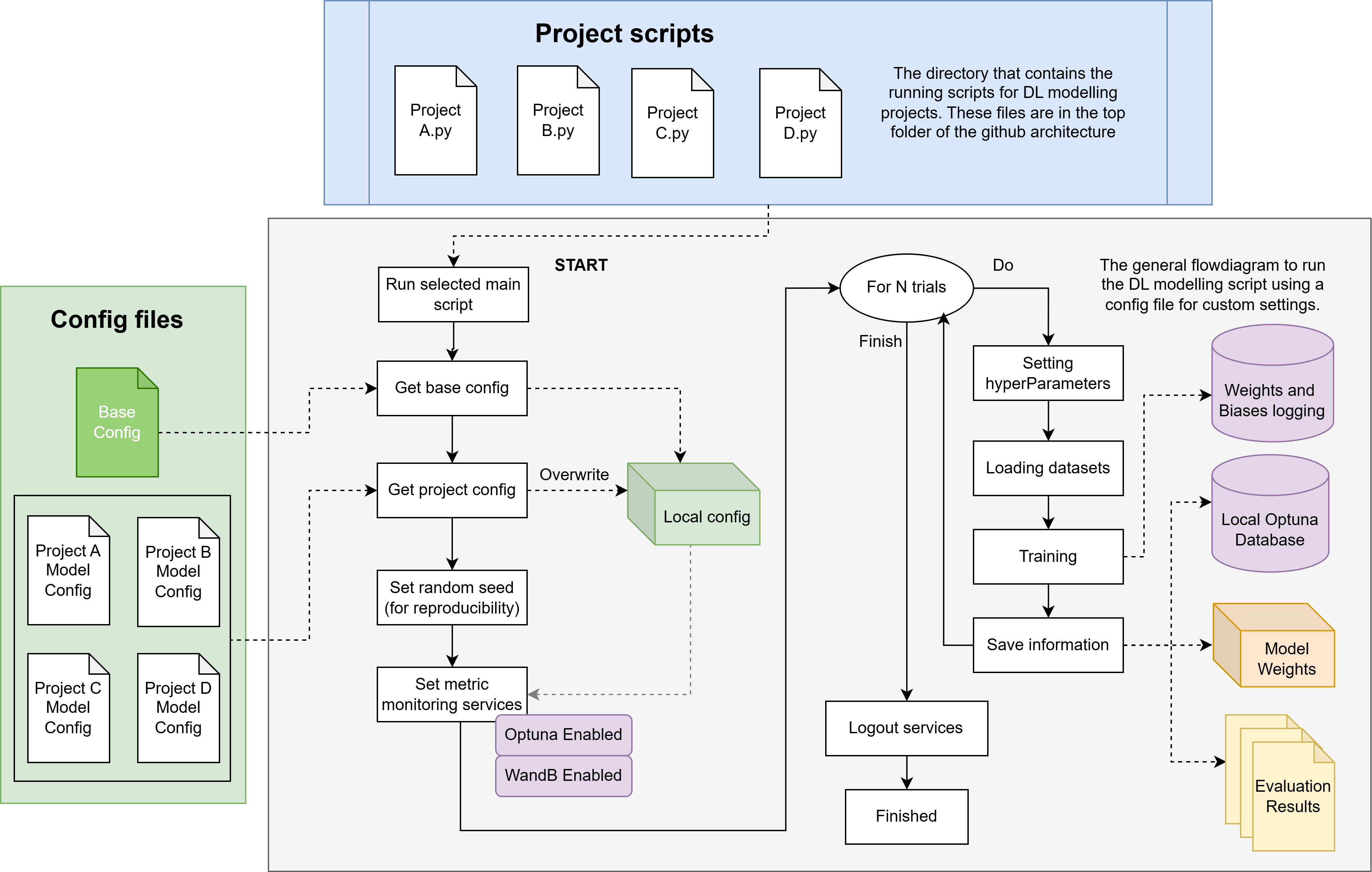}
    \caption{Code flow diagram of the PR3DICTR framework.}
    \label{fig:code_flow}
\end{figure}

Within the configuration system, PR3DICTR employs a Base Config (Figure~\ref{fig:code_flow}), which provides default values for all possible parameters and acts as a general template. This ensures that users do not need to redefine every parameter for each new project, needing only to make a project-specific config. For example, a user who wishes to change only the model architecture can simply define that parameter in the project config, while the preprocessing and augmentation settings remain as specified in the Base Config.

\subsection{The PR3DICTR architecture}

\begin{figure}[H]
    \centering
    \includegraphics[width=0.9\textwidth]{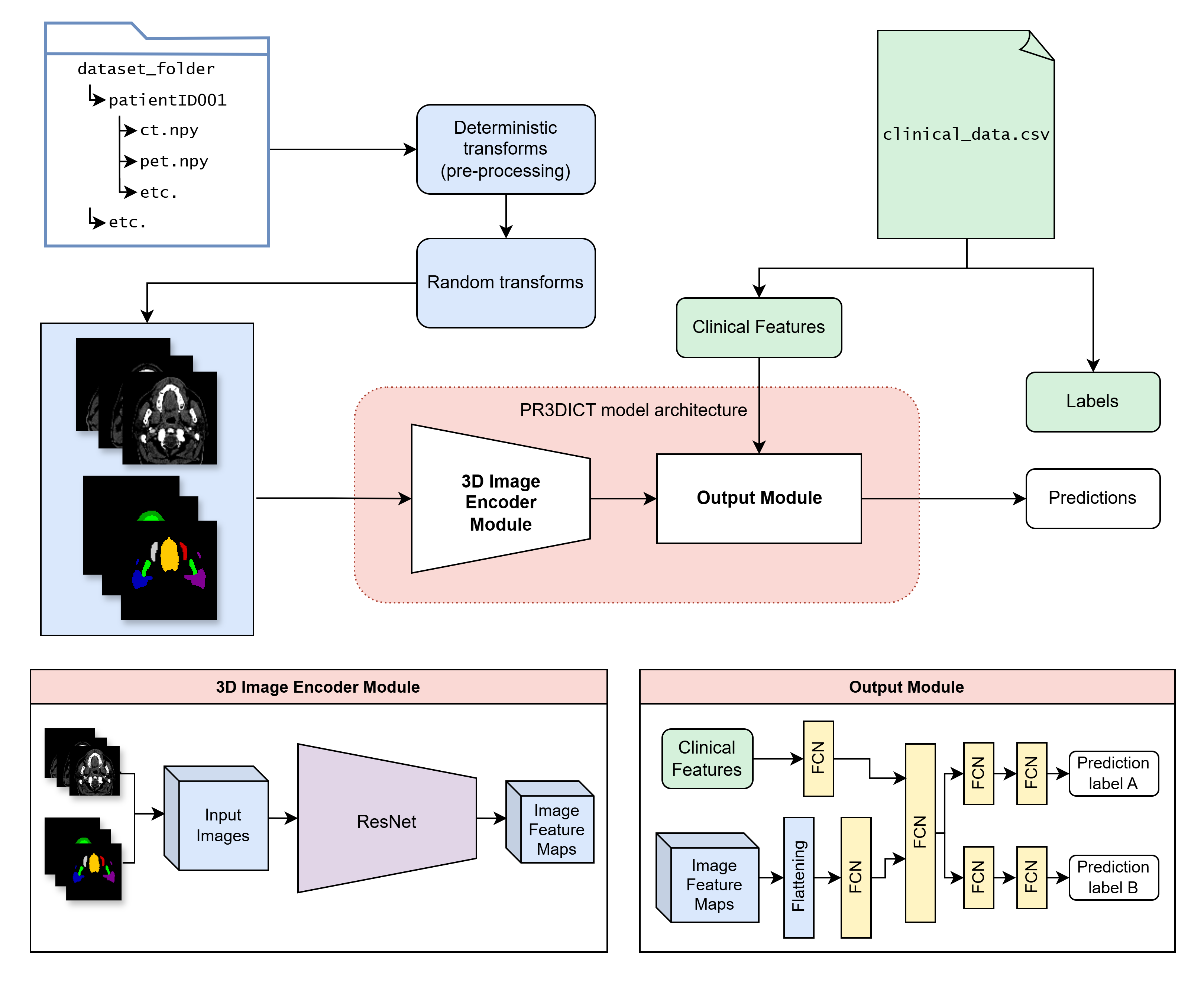}
    \caption{Outline of the classification model framework in PR3DICTR, for a hypothetical 2-label classification task.}
    \label{fig:architecture}
\end{figure}

Models trained within the PR3DICTR framework consist of two main modules: an image encoder and an output module (Figure~\ref{fig:architecture}). The image encoder serves as the backbone of the model, responsible for extracting image features from the input data. Through the configuration file, users can select the encoder architecture, such as a ResNet or DenseNet, from the currently implemented options (Table~\ref{tab:options}). The encoder processes all input image modalities simultaneously. When multiple 3D modalities are provided (e.g., CT and PET), they are stacked along the channel dimension. Consequently, the model input tensor has the shape $[B, C, H, W, D]$, where $B$ denotes the batch size, $C$ the number of channels (equal to the number of input modalities per patient), and $H$, $W$, and $D$ the height, width, and depth of the images, respectively.

The resulting feature map is then passed to the output module, which integrates the extracted image features with clinical (tabular) data via fully connected linear layers or by a ViT. This module is also highly configurable through the config file---for example, users can specify the number of linear layers, the layer at which clinical features are concatenated, and the dropout rate. The output module further supports multi-label classification. As demonstrated with two classes in Figure~\ref{fig:architecture}, for each label (class) defined in the config, a corresponding output head is created. Each output head can comprise an independent set of linear layers, enabling label-specific representations that are not shared across outputs.

When only clinical features serve as input, model training can be performed using a multilayer perceptron (MLP). Hence, no image types should be specified in the config. Subsequently, the image feature encoder module will be empty. The MLP inherits the settings of the output module, except that the layers used to process image features are removed, enabling straightforward training with tabular inputs only.

\small
\begin{longtable}{P{0.16\textwidth}P{0.18\textwidth}P{0.58\textwidth}}
\caption{Main options within the PR3DICTR framework modules.}\label{tab:options}\\
\toprule
\textbf{Type} & \textbf{Options} & \textbf{Notes} \\
\midrule
\endfirsthead
\toprule
\textbf{Type} & \textbf{Options} & \textbf{Notes} \\
\midrule
\endhead
\midrule
\multicolumn{3}{r}{Continued on next page} \\
\endfoot
\bottomrule
\endlastfoot

Image modalities & CT & Standard planning or diagnostic CT used as anatomical input. \\
 & PET & Functional metabolic imaging; often co-registered with CT. \\
 & Radiation dose & 3D dose distribution map, used in radiotherapy settings. \\
 & Segmentations & Binary or multi-class masks for organs or targets; used as additional spatial priors. \\
 & MRI & Multi-sequence (T1, T2, etc.) structural or functional MRI. \\
 & Other & Any additional volumetric image (e.g. perfusion, ADC, or synthetic images) can be added to the PR3DICTR framework. \\
\hline
Dataset types & Standard & Loads data directly from the disk on each epoch. Slowest for model training but consumes the least memory. \\
 & Cache & Caches pre-processed (all non-random transforms) data in memory for faster training. Reduces epoch loading time but requires sufficient RAM. \\
 & SmartCache & A user-defined fraction of the dataset is cached in RAM rather than the whole dataset, balancing speed and memory usage. \\
 & Persistent & Similar to CacheDataset, but caches deterministically transformed data on disk rather than in RAM. \\
\hline
Input preprocessing & Value clipping and normalisation & Clips the values of the image inputs to a range $[a_{\min}, a_{\max}]$ and normalises it to $[b_{\min}, b_{\max}]$. Example: CT clipped to [-200, 400] HU and normalised to [0,1]. \\
 & Value mapping & Intended for categorical structures (e.g. organ contours). Allows selected structures to be assigned specific values while all others are set to 0. \\
 & Cropping & Crops the image, using the centre point of the volume, to a fixed size. \\
\hline
Random transforms & Cropping & Randomly crops the images to a fixed size, using different centre-points. \\
 & Flipping & Flips the 3D images horizontally (X axis), with a fixed probability of 50\%. \\
 & Affine transformation & Includes translation, scaling, and shear; preserves spatial relationships. \\
 & Rotation & Random 3D rotation around multiple axes. \\
 & Noise & Adds Gaussian noise to simulate acquisition variability. \\
 & MixUp & Uses the MixUp algorithm \cite{Zhang2018} to mix input-label pairs during training to improve generalisability and calibration. \\
\hline
Label types & Binary & Standard binary classification (e.g., yes/no). \\
 & Event & Time-to-event or survival endpoints. Requires two label columns, e.g. \texttt{X\_event} and \texttt{X\_\{unit\}}. \\
\hline
Models & CNN & Generic convolutional backbone for volumetric data. Number of layers and feature maps can be defined in the config. \\
 & ResNet & Residual network for efficient gradient flow \cite{He2015}. Sizes: ResNet-10, 18, 34, 50, 101, 152, and 200. \\
 & DenseNet & Dense connectivity for compact yet expressive feature extraction \cite{Huang2018}. Sizes: 121, 169, 201, and 264. \\
 & EfficientNetV2 & Compact, high-performance CNN architecture using progressive scaling and fused convolutions \cite{Tan2021}. Variants: XS, S, M, L, XL. \\
 & ConvNeXt & Modernised CNN with transformer-like design. Supports 3D ConvNeXt tiny, small, base, large, and xlarge. \\
 & ViT & Vision Transformer operating directly on image patches \cite{Dosovitskiy2021}. \\
 & TransRP & Hybrid CNN-ViT architecture where the CNN can be any of the other CNN-based architectures \cite{Ma2023}. \\
 & MLP & Multilayer perceptron using only clinical features and no imaging features. \\
\hline
Output module & N shared layers & Number of shared linear layers following the flattening layer. \\
 & N endpoint layers & Number of non-shared linear layers per output head. \\
 & N clinical layers & Number of linear layers applied to clinical features before concatenation. \\
 & Clinical layers position & Index of the shared linear layer at which clinical features are concatenated. \\
 & Linear layer sizes & Size of each shared, endpoint, and clinical layer. \\
\hline
Loss functions & BCE & Binary cross-entropy for standard classification. \\
 & Focal & Down-weights easy examples; handles class imbalance. \\
 & Hill & Hill loss for smooth probabilistic calibration in imbalanced settings. \\
 & ASL & Asymmetric loss variant improving recall on minority classes. \\
 & NLL & Negative log-likelihood for survival/time-to-event models. \\
\hline
Optimisers & Adam & Adaptive moment estimation; strong general-purpose choice \cite{Kingma2017}. \\
 & AdamW & Adam with decoupled weight decay for better regularisation \cite{Loshchilov2019}. \\
 & AdaBound & Transitions from adaptive to SGD behaviour during training \cite{Luo2019}. \\
 & SGD & Stochastic gradient descent; strong baseline with momentum. \\
\hline
Scheduler & Cosine & Cosine annealing schedule for smooth learning-rate decay. \\
 & Step & Reduces the learning rate by a fixed factor at predefined intervals. \\
 & Plateau & Lowers the learning rate when a monitored metric stops improving. \\
 & None & Uses a fixed learning rate throughout training. \\
\hline
Evaluation metrics & AUC & Area under the ROC curve; threshold-independent classifier performance. \\
 & Accuracy & Fraction of correct predictions at a fixed threshold or a Youden-J-derived threshold. \\
 & C-index & Concordance index for survival/time-to-event predictions. \\
 & F1-score & Harmonic mean of precision and recall. \\
 & Precision & Proportion of predicted positives that are true. \\
 & Recall & Proportion of true positives correctly predicted. \\
 & ACE & Adaptive calibration error \cite{Nixon2020}. \\
 & ECE & Expected calibration error \cite{Nixon2020}. \\
 & MCE & Maximum calibration error \cite{Nixon2020}. \\
 & Brier score & Mean squared error of probabilistic predictions. \\
\hline
Visualisations & Calibration plot & Plots predicted vs. observed probabilities for equal-frequency bins. \\
 & Reliability plot & Similar to calibration plots, but uses fixed-width bins. \\
 & Confusion matrix & Summarises TP, FP, FN, and TN counts. \\
 & ROC curve & Plots TPR vs. FPR across thresholds. \\
 & Kaplan--Meier curve & Visualises survival probabilities across groups for event-based endpoints. \\

\end{longtable}

\normalsize

\subsection{Dataset, loaders and augmentation}

A key component of the PR3DICTR framework is the data handling module, which manages dataset loading and data augmentation.

Image data augmentation in PR3DICTR occurs in two stages: deterministic and non-deterministic transforms. Deterministic transforms include all input preprocessing operations required to convert image files into model-ready inputs, such as cropping, windowing and scaling, the latter two of which can be specified per input channel (e.g. HU range for CT scans or SUV for PET scans). For segmentation data it is also possible to remap class values, for which a required intensity per segmentation needs to be given. This can be used to give more weight to certain structures. These deterministic transforms are applied consistently across all dataset splits. Non-deterministic transforms, by contrast, introduce stochastic variation and are typically applied only to the training set. They include operations such as random flipping, rotation, and the MixUp algorithm (see Table~\ref{tab:options} for all options).

Based on the configuration settings, the framework constructs a dataloader using one of four available MONAI Dataset interfaces, selected according to the user's computational and hardware requirements (Table~\ref{tab:options}). For example, the CacheDataset typically results in the shortest training epoch times but may require a very high amount of RAM when the training set consists of many large 3D images. In addition, the PR3DICTR dataloader includes a convenient \texttt{get\_patient()} function, enabling retrieval of a specific patient's data via their patient ID. This functionality is particularly useful for post hoc analyses and debugging.

\subsection{Training, hyperparameter optimization and experiments}

As indicated in Figure~\ref{fig:overall_workflow} (step 5), model training in PR3DICTR can be performed in two modes: a standard training mode or an experiment-based optimization mode. In the standard mode, a single K-fold cross-validation is conducted using the hyperparameters defined in the configuration file. The number of folds ($K$) and the clinical variables used for stratified sampling are both configurable, ensuring balanced data splits across folds. This approach provides robust model evaluation while maintaining flexibility for different dataset characteristics and project requirements. For computational efficiency during the modelling exploration phase, users may also choose to train on only a subset of the generated folds---for example, running models on two or three of five folds---while still maintaining consistent fold definitions and reproducibility.

In the experiment-based mode, PR3DICTR integrates Optuna to perform automated hyperparameter optimization \cite{Akiba2019}. Optuna provides a search algorithm to automatically conduct hyperparameter space optimization, leading to converging optimisation rather than extensive grid search. Each experiment corresponds to one Optuna optimization run and consists of multiple trials, with each trial representing a unique combination of hyperparameters proposed by the search algorithm. Within each trial, PR3DICTR performs a complete K-fold cross-validation, identical in structure to the standard mode. The performance metrics from all folds are then aggregated to compute the trial's objective value, which Optuna uses to guide subsequent trials toward improved hyperparameter configurations.

\subsection{Model training evaluation}

For each fold of every trial within an experiment setup, a set of evaluation and logging files are stored. This includes a copy of the config, saved as a YAML file which enables quick recall of the settings used for the given model instance, as well as the model weights (optional), model predictions on all patients, and CSV files and plots for the model evaluation on the training and validation cohorts.

A set of standardised evaluation metrics and visualisation techniques are built into the PR3DICTR framework. These include a mix of classification and calibration metrics for both binary and event-based labels, as well as different visualisations of model performance. These metrics and visualisations are listed in Table~\ref{tab:options} and are computed or plotted for each label class separately.

During training, Weights \& Biases (W\&B) integration provides automatic experiment tracking and visualization. Each fold within each trial is logged independently, including the loss and metrics on the training and validation set for each epoch. This enables detailed performance monitoring across folds, trials, and experiments, facilitating transparent comparisons and reproducible research.

\subsection{Model evaluation on the test set}

For final evaluation on the test set (Figure~\ref{fig:overall_workflow}, step 6), the PR3DICTR framework incorporates a post-hoc evaluation function, which retrieves the model weights from the folders saved during training, and runs a pass over the test set. The results on the test set are saved using the same evaluation metrics and visualisations as the train and validation set and are saved per fold (i.e. per model) as well as for the ensemble predictions. The choice to design the evaluation function in a post-hoc manner also facilitates simplified validation of final or existing deep learning models on external test sets; the same function can be applied over datasets from multiple different centres.

\section{Example use case}

As an example, the PR3DICTR framework was trained and evaluated on a sex classification task on a newly retrieved dataset from The Cancer Imaging Archive: the NSCLC-Radiomics database by Aerts et al. \cite{Aerts2014}. Two example Jupyter notebooks are provided: one introduces the data preprocessing required (Figure~\ref{fig:overall_workflow}, steps 1--3), and the other walks users through developing a model using the PR3DICTR framework (Figure~\ref{fig:overall_workflow}, steps 4--6). The data consisted of thorax CT scans, segmentation masks of the lungs, and a CSV containing tabular data. The model used was the default ResNet model (ResNet-10) implemented in PR3DICTR and was able to produce a nearly perfect distinction between the sexes in the test set, with decent calibration. The example notebooks can be found at \url{https://github.com/DLinRadiotherapyUMCG/PR3DICTR/tree/main/notebooks/01_LearningExamples}.

\begin{figure}[htbp]
    \centering
    \hfill
    \begin{subfigure}[b]{0.4\textwidth}
        \includegraphics[width=\textwidth]{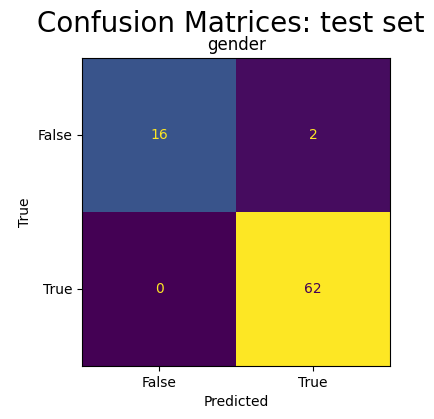}
    \end{subfigure}
    \hfill
    \begin{subfigure}[b]{0.38\textwidth}
        \includegraphics[width=\textwidth]{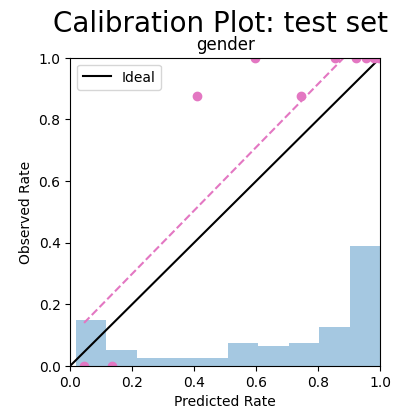}
    \end{subfigure}
    \hfill
    \hfill
    \caption{Confusion matrix and calibration plot for the sex prediction model produced using the PR3DICTR framework on the NSCLC-Radiomics database.}
    \label{fig:example_results}
\end{figure}

\section{Discussion}

The PR3DICTR framework has been developed to help streamline and standardize model development across a wide range of prediction and classification tasks based on 3D medical imaging, while being built with a highly modular design. By utilizing a user-adjustable configuration file, the framework enables straightforward customization of model parameters and hyperparameter tuning to suit specific tasks. Comprehensive guidelines for data preparation are provided to ensure compatibility and ease of use. A diverse set of model architectures, training, and evaluation options have been integrated into the modular design, allowing each model to be tailored precisely to its intended application. This flexibility positions the PR3DICTR framework as a robust and adaptable tool for research in deep learning classification applications using 3D medical imaging.

Simplification, accessible usability, and adaptability were central objectives in the development of the PR3DICTR framework. The framework eliminates the need to re-implement models from scratch for each new experiment, maintaining a consistent structure while allowing flexible modifications. Irrelevant components are abstracted away from the user, while configurable elements are integrated into configuration files. This design enables users with theoretical understanding to select options---such as model architectures, loss functions or optimization strategies---without needing to implement them manually, while still allowing for the option of adaptation when desired. Best performing models are automatically saved, and key performance metrics are readily accessible via structured CSV files and visualizations. Integration with Weights \& Biases facilitates experiment tracking and real-time monitoring of model training. Additionally, the inclusion of Optuna enables automated hyperparameter optimization within user-specified ranges, further streamlining the model development process. Collectively, these features contribute to a highly accessible and efficient workflow.

Operational standardization was another key aim of our framework, aimed at promoting transparency and repeatability in scientific research. By employing a consistent pipeline across experiments and potentially across related research projects within an organization, the framework ensures that methodological choices are clearly documented and easily traceable. Each trained model is accompanied by a saved configuration file, enabling transparency to replicate model training by the original researcher, collaborators, or external reviewers if needed. Evaluation metrics are automatically computed upon model completion and stored in a standardized format, facilitating reliable and straightforward comparisons across experiments. This emphasis on reproducibility and consistency strengthens the framework's utility in rigorous research environments.

Modularity is the final foundational principle addressed by the PR3DICTR framework. The framework is organized into separate, interchangeable modules, enabling broad applicability across diverse tasks. Users can select which modules to include via a configuration file. For example, they can specify multiple input channels to enable multi-modal modelling with CT and PET scans or restrict the model to a single modality if desired. Different model architectures are also available, allowing both convolutional and transformer-based architectures to be integrated seamlessly. Furthermore, the PR3DICTR framework facilitates the fusion of tabular data within image models, further enhancing the multi-modality of trained models, and the ability to train non-image models (i.e. MLPs). Unlike frameworks such as MONAI, data loaders and training functions are pre-implemented, requiring users only to adjust key parameters, such as the number of cross-validation folds or epochs for early stopping. Model evaluation is similarly flexible, with a comprehensive set of methods available so users can choose the most appropriate metrics and visualisations for their application. Together, these features highlight the versatility and modularity of the PR3DICTR framework, making it adaptable to a wide range of modelling scenarios.

The PR3DICTR framework has been the result of our research focuses on developing deep learning–based normal tissue complication probability (NTCP) and outcome prediction models for radiotherapy using three-dimensional data, including 3D CT images, radiation dose distributions, and organ-at-risk contours. Within this setting, the PR3DICTR framework has been built towards standardization to improve the consistency and continency between projects. Prior to its adoption, our research group relied on multiple separate—albeit conceptually similar—implementations of deep learning (DL) NTCP model training code, for example in the development of DL xerostomia \cite{Chu2025}  and DL dysphagia \cite{deVette2025} models. By adopting PR3DICTR as the basis for our ongoing work, we have been able to standardise model development pipelines across a wide range of applications, thereby reducing implementation-driven variability and facilitating more reliable comparisons between models, endpoints, and disease sites. This standardisation has enabled, for instance, the use of identical data-loading, training, and evaluation code across head and neck as well as lung cancer datasets, requiring only minimal changes to model configuration files. The config-driven design further improves transparency and reproducibility – for the researcher themselves as between researchers – by explicitly documenting all experiment-specific choices and enabling straightforward replication of models across projects. Moreover, the flexibility of these configuration files allows seamless switching between binary classification tasks, such as xerostomia \cite{Chu2025} and dysphagia \cite{deVette2025} prediction, and multi-endpoint models, including a multi-toxicity NTCP model \cite{MacRae2026}, while retaining the same core training and evaluation framework. As a result, methodological developments—such as architectural modifications, loss functions, or training strategies—can be rapidly propagated across projects without reimplementation. Finally, the modular design of PR3DICTR has provided a robust foundation for extending the framework with case-specific components, such as the incorporation of uncertainty quantification methods for DL NTCP models \cite{MacRae2025}, implemented as extensions to the core pipeline rather than standalone codebases. While the initial effort required to refactor legacy implementations into a unified framework was non-trivial, this was offset by substantial gains in maintainability, extensibility, and scalability, demonstrating that PR3DICTOR serves not only as a unifying framework for existing DL NTCP models but also as a foundation for future methodological development.

While the proposed framework introduces a modular and standardized approach to deep learning model development, certain challenges remain unaddressed. One notable limitation is the exact reproducibility of results. Although reproducibility was prioritized by saving detailed configuration files that include architecture specifications and hyperparameters for each model, minor discrepancies may still occur due to differences in hardware and floating-point rounding errors. These factors can lead to slight deviations in outcomes even when models are trained using identical configurations. Importantly, not all steps in the modelling pipeline can or should be fully standardised. In particular, pre-processing and post-processing often involve data-specific decisions that depend on the dataset, imaging protocol, endpoint, and clinical context. Although some of these choices may be partly arbitrary, they remain an important part of the modelling process and require careful consideration by the user. PR3DICTR standardises the overall framework and implementation structure, but does not eliminate the need for critical judgement in dataset-specific methodological choices. Furthermore, this framework does not aim to redefine the process or accessibility of DL model creation. Instead, it offers a streamlined and flexible structure that promotes consistency and reusability. We acknowledge that user preferences vary; some may favour highly abstracted, low-code environments, while others prefer granular control through code. Nevertheless, we believe this framework strikes a balance that appeals to a broad range of researchers, particularly those with a solid understanding of DL principles, who seek to accelerate model development without sacrificing transparency or flexibility.

Additional features are currently under development to enhance the framework's capabilities. Multi-class classification can enable more complex prediction tasks and could be implemented as an extension of the existing single-class classification options, leveraging the modular architecture of the framework. Emerging research areas, such as model uncertainty quantification, will also be integrated in the near future. In clinical settings, quantifying uncertainty can support more informed decision-making and facilitate performance monitoring. Furthermore, visualization techniques, including attention maps, will be incorporated to improve interpretability and model transparency. Finally, we would like to extend reproducibility and standardisation through semi-automatic generation of model cards \cite{Barragn-Montero2025}, which report the primary information related to the training of models developed within PR3DICTR (e.g. hyperparameters, cohort sizes, etc.). Further potential additions include GUI components, including an interface to easily set up the config files. These advancements aim to strengthen the framework's applicability across diverse domains and promote more robust, explainable AI solutions.

\section{Conclusion}

The PR3DICTR-framework offers a modular, standardized, and user-friendly solution for developing 3D medical image-based deep learning models in prediction, detection or other classification tasks, streamlining both experimentation and reproducibility. Its design balances flexibility with simplicity, enabling researchers to efficiently build and evaluate models without sacrificing transparency or control. Ongoing enhancements, including support for multi-class classification, uncertainty estimation, and interpretability tools, aim to further expand its utility across diverse research and clinical applications.

\section*{Acknowledgements}
Research supported by the NVIDIA Academic Grant Program using DGX-Spark. 




\bibliographystyle{IEEEtran}
\bibliography{pr3dictr_bib}

@article{MacRae2025,
   author = {DC MacRae and L van der Hoek and JE van Aalst and SPM de Vette and R van der Wal and Hendrike Neh and Baoqiang Ma and NM Sijtsema and MA Valdenegro-Toro and PMA van Ooijen and LV van Dijk},
   month = {12},
   title = {An Evaluation of Uncertainty Quantification Methods and Measures for Deep Learning Outcome Prediction Models in Head and Neck Cancer Radiotherapy},
   url = {https://ssrn.com/abstract=6041252},
   year = {2025}
}

@article{MacRae2026,
   author = {DC MacRae and L van der Hoek and SPM de Vette and H Neh and AC Moreno and CD Fuller and JA Langendijk and MA Valdenegro-Toro and NM Sijtsema and PMA van Ooijen and LV van Dijk},
   doi = {10.1016/j.radonc.2026.111486},
   issn = {0167-8140},
   journal = {Radiotherapy and Oncology},
   month = {3},
   publisher = {Elsevier},
   title = {A multi-toxicity deep learning approach for normal tissue complication probability modelling in head and neck cancer patients receiving radiotherapy},
   url = {https://doi.org/10.1016/j.radonc.2026.111486},
   year = {2026}
}

@article{deVette2025,
   author = {S P M de Vette and H Neh and L Van Der Hoek and D C MacRae and H Chu and A Gawryszuk and R.J.H.M. Steenbakkers and P M A Van Ooijen and C D Fuller and K A Hutcheson and J A Langendijk and N M Sijtsema and L V Van Dijk},
   doi = {10.1016/j.radonc.2025.111169},
   issn = {01678140},
   journal = {Radiotherapy and Oncology},
   month = {12},
   pages = {111169},
   title = {Deep Learning NTCP Model for Late Dysphagia after Radiotherapy for Head and Neck Cancer Patients Based on 3D Dose, CT and Segmentations},
   volume = {213},
   year = {2025}
}

@article{Chu2025,
   author = {Hung Chu and Suzanne P M de Vette and Hendrike Neh and Nanna M Sijtsema and Roel J H M Steenbakkers and Amy Moreno and Johannes A Langendijk and Peter M A van Ooijen and Clifton D Fuller and Lisanne V Van Dijk},
   doi = {10.1016/j.ijrobp.2024.07.2334},
   issn = {0360-3016},
   issue = {1},
   journal = {International Journal of Radiation Oncology, Biology, Physics},
   month = {1},
   pages = {269-280},
   publisher = {Elsevier},
   title = {Three-Dimensional Deep Learning Normal Tissue Complication Probability Model to Predict Late Xerostomia in Patients With Head and Neck Cancer},
   volume = {121},
   year = {2025}
}

@article{Akiba2019,
   author = {Takuya Akiba and Shotaro Sano and Toshihiko Yanase and Takeru Ohta and Masanori Koyama},
   month = {7},
   title = {Optuna: A Next-generation Hyperparameter Optimization Framework},
   url = {http://arxiv.org/abs/1907.10902},
   year = {2019}
}

@article{Nixon2020,
   author = {Jeremy Nixon and Mike Dusenberry and Ghassen Jerfel and Timothy Nguyen and Jeremiah Liu and Linchuan Zhang and Dustin Tran},
   month = {8},
   title = {Measuring Calibration in Deep Learning},
   url = {http://arxiv.org/abs/1904.01685},
   year = {2020}
}

@article{Kingma2017,
   author = {Diederik P. Kingma and Jimmy Ba},
   month = {1},
   title = {Adam: A Method for Stochastic Optimization},
   url = {http://arxiv.org/abs/1412.6980},
   year = {2017}
}

@article{Loshchilov2019,
   author = {Ilya Loshchilov and Frank Hutter},
   month = {1},
   title = {Decoupled Weight Decay Regularization},
   url = {http://arxiv.org/abs/1711.05101},
   year = {2019}
}

@article{Luo2019,
   author = {Liangchen Luo and Yuanhao Xiong and Yan Liu and Xu Sun},
   month = {2},
   title = {Adaptive Gradient Methods with Dynamic Bound of Learning Rate},
   url = {http://arxiv.org/abs/1902.09843},
   year = {2019}
}

@inproceedings{Ma2023,
   author = {Baoqiang Ma and Jiapan Guo and Lisanne V. van Dijk and Peter M.A. van Ooijen and Stefan Both and Nanna M. Sijtsema},
   booktitle = {Medical Imaging and Deep Learning},
   title = {TransRP: Transformer-based PET/CT feature extraction incorporating clinical data for recurrence-free survival prediction in oropharyngeal cancer},
   year = {2023}
}

@article{Dosovitskiy2021,
   author = {Alexey Dosovitskiy and Lucas Beyer and Alexander Kolesnikov and Dirk Weissenborn and Xiaohua Zhai and Thomas Unterthiner and Mostafa Dehghani and Matthias Minderer and Georg Heigold and Sylvain Gelly and Jakob Uszkoreit and Neil Houlsby},
   month = {6},
   title = {An Image is Worth 16x16 Words: Transformers for Image Recognition at Scale},
   url = {http://arxiv.org/abs/2010.11929},
   year = {2021}
}

@article{Tan2021,
   author = {Mingxing Tan and Quoc V. Le},
   month = {6},
   title = {EfficientNetV2: Smaller Models and Faster Training},
   url = {http://arxiv.org/abs/2104.00298},
   year = {2021}
}

@article{Huang2018,
   author = {Gao Huang and Zhuang Liu and Laurens van der Maaten and Kilian Q. Weinberger},
   month = {1},
   title = {Densely Connected Convolutional Networks},
   url = {http://arxiv.org/abs/1608.06993},
   year = {2018}
}

@article{He2015,
   author = {Kaiming He and Xiangyu Zhang and Shaoqing Ren and Jian Sun},
   month = {12},
   title = {Deep Residual Learning for Image Recognition},
   url = {http://arxiv.org/abs/1512.03385},
   year = {2015}
}

@article{Zhang2018,
   author = {Hongyi Zhang and Moustapha Cisse and Yann N. Dauphin and David Lopez-Paz},
   month = {4},
   title = {mixup: Beyond Empirical Risk Minimization},
   url = {http://arxiv.org/abs/1710.09412},
   year = {2018}
}

@misc{Barragn-Montero2025,
   author = {Ana M. Barragán-Montero and Margerie Huet-Dastarac and Silvia M. Herranz-Hernández and Benjamin Tengler and Emma Skarsø Buhl and Arthur Galapon and Carlos E. Cárdenas and Marco Fusella and Geoffroy Herbin and Yvonne de Hond and Franziska Knuth and Ciaran Malone and Peter van Ooijen and Charlotte Robert and Michele Zeverino and Coen Hurkmans and Tomas Janssen and Stine Sofia Korreman and Charlotte L. Brouwer},
   doi = {https://doi.org/10.5281/zenodo.17399354},
   publisher = {Zenodo},
   title = {AID-RT: Standardising AI Documentation in RadioTherapy with a domain-specific model card.},
   url = {https://zenodo.org/records/17399354},
   year = {2025}
}

@article{Paszke2019,
   author = {Adam Paszke and Sam Gross and Francisco Massa and Adam Lerer and James Bradbury and Gregory Chanan and Trevor Killeen and Zeming Lin and Natalia Gimelshein and Luca Antiga and Alban Desmaison and Andreas Köpf and Edward Yang and Zach DeVito and Martin Raison and Alykhan Tejani and Sasank Chilamkurthy and Benoit Steiner and Lu Fang and Junjie Bai and Soumith Chintala},
   doi = {https://doi.org/10.48550/arXiv.1912.01703},
   month = {12},
   title = {PyTorch: An Imperative Style, High-Performance Deep Learning Library},
   url = {http://arxiv.org/abs/1912.01703},
   year = {2019}
}

@article{Pati2023,
   author = {Sarthak Pati and Siddhesh P. Thakur and İbrahim Ethem Hamamcı and Ujjwal Baid and Bhakti Baheti and Megh Bhalerao and Orhun Güley and Sofia Mouchtaris and David Lang and Spyridon Thermos and Karol Gotkowski and Camila González and Caleb Grenko and Alexander Getka and Brandon Edwards and Micah Sheller and Junwen Wu and Deepthi Karkada and Ravi Panchumarthy and Vinayak Ahluwalia and Chunrui Zou and Vishnu Bashyam and Yuemeng Li and Babak Haghighi and Rhea Chitalia and Shahira Abousamra and Tahsin M. Kurc and Aimilia Gastounioti and Sezgin Er and Mark Bergman and Joel H. Saltz and Yong Fan and Prashant Shah and Anirban Mukhopadhyay and Sotirios A. Tsaftaris and Bjoern Menze and Christos Davatzikos and Despina Kontos and Alexandros Karargyris and Renato Umeton and Peter Mattson and Spyridon Bakas},
   doi = {10.1038/s44172-023-00066-3},
   issn = {2731-3395},
   issue = {1},
   journal = {Communications Engineering},
   month = {5},
   pages = {23},
   title = {GaNDLF: the generally nuanced deep learning framework for scalable end-to-end clinical workflows},
   volume = {2},
   url = {https://www.nature.com/articles/s44172-023-00066-3},
   year = {2023}
}

@article{Wehbe2024,
   author = {Alaa Wehbe and Silvana Dellepiane and Irene Minetti},
   doi = {10.1109/ACCESS.2024.3462629},
   issn = {2169-3536},
   journal = {IEEE Access},
   pages = {141414-141424},
   title = {Enhanced Lung Cancer Detection and TNM Staging Using YOLOv8 and TNMClassifier: An Integrated Deep Learning Approach for CT Imaging},
   volume = {12},
   url = {https://ieeexplore.ieee.org/document/10681569/},
   year = {2024}
}

@article{Erasmus2023,
   author = {Lauren T. Erasmus and Taylor A. Strange and Rishi Agrawal and Chad D. Strange and Jitesh Ahuja and Girish S. Shroff and Mylene T. Truong},
   doi = {10.3390/diagnostics13213359},
   issn = {2075-4418},
   issue = {21},
   journal = {Diagnostics},
   month = {11},
   pages = {3359},
   title = {Lung Cancer Staging: Imaging and Potential Pitfalls},
   volume = {13},
   url = {https://www.mdpi.com/2075-4418/13/21/3359},
   year = {2023}
}

@article{vanGalen2025,
   author = {Isa F. van Galen and Camila R. Guetter and Elisa Caron and Jeremy Darling and Jemin Park and Roger B. Davis and Mikayla Kricfalusi and Virendra I. Patel and Joost A. van Herwaarden and Thomas F.X. O'Donnell and Marc L. Schermerhorn},
   doi = {10.1016/j.jvs.2024.12.129},
   issn = {07415214},
   issue = {5},
   journal = {Journal of Vascular Surgery},
   month = {5},
   pages = {1023-1032.e1},
   title = {The effect of aneurysm diameter on perioperative outcomes following complex endovascular repair},
   volume = {81},
   url = {https://linkinghub.elsevier.com/retrieve/pii/S0741521425000163},
   year = {2025}
}

@article{Huang2024,
   author = {Legang Huang and Jiankuan Lu and Ying Xiao and Xiaofei Zhang and Cong Li and Guangchao Yang and Xiangfei Jiao and Zijie Wang},
   doi = {10.3389/fcvm.2024.1354517},
   issn = {2297-055X},
   journal = {Frontiers in Cardiovascular Medicine},
   month = {2},
   title = {Deep learning techniques for imaging diagnosis and treatment of aortic aneurysm},
   volume = {11},
   url = {https://www.frontiersin.org/articles/10.3389/fcvm.2024.1354517/full},
   year = {2024}
}

@article{Saba2025,
   author = {Luca Saba and Ernesto D’Aloja},
   doi = {10.1038/s41746-025-01791-z},
   issn = {2398-6352},
   issue = {1},
   journal = {npj Digital Medicine},
   month = {7},
   pages = {392},
   title = {Predictive techniques in medical imaging: opportunities, limitations, and ethical-economic challenges},
   volume = {8},
   url = {https://www.nature.com/articles/s41746-025-01791-z},
   year = {2025}
}

@article{Vagvala2022,
   author = {Saivenkat Vagvala and Jeffrey P. Guenette and Camilo Jaimes and Raymond Y. Huang},
   doi = {10.1186/s40644-022-00455-5},
   issn = {1470-7330},
   issue = {1},
   journal = {Cancer Imaging},
   month = {12},
   pages = {19},
   title = {Imaging diagnosis and treatment selection for brain tumors in the era of molecular therapeutics},
   volume = {22},
   url = {https://cancerimagingjournal.biomedcentral.com/articles/10.1186/s40644-022-00455-5},
   year = {2022}
}

@article{Rong2024,
   author = {Jian Rong and Yutong Liu},
   doi = {10.1186/s44330-024-00010-7},
   issn = {3004-8729},
   issue = {1},
   journal = {BMC Methods},
   month = {8},
   pages = {10},
   title = {Advances in medical imaging techniques},
   volume = {1},
   url = {https://bmcmethods.biomedcentral.com/articles/10.1186/s44330-024-00010-7},
   year = {2024}
}

@article{Aiello2019,
   author = {Marco Aiello and Carlo Cavaliere and Antonio D’Albore and Marco Salvatore},
   doi = {10.3390/jcm8030316},
   issn = {2077-0383},
   issue = {3},
   journal = {Journal of Clinical Medicine},
   month = {3},
   pages = {316},
   title = {The Challenges of Diagnostic Imaging in the Era of Big Data},
   volume = {8},
   url = {https://www.mdpi.com/2077-0383/8/3/316},
   year = {2019}
}

@misc{Aerts2014,
   author = {H. J. W. L. Aerts and L. Wee and E. Rios Velazquez and R. T. H. Leijenaar and C. Parmar and P. Grossmann and S. Carvalho and J. Bussink and R. Monshouwer and B. Haibe-Kains and D. Rietveld and F. Hoebers and M. M. Rietbergen and C. R. Leemans and A. Dekker and J. Quackenbush and R. J. Gillies and P. Lambin},
   doi = {https://doi.org/10.7937/K9/TCIA.2015.PF0M9REI},
   publisher = {The Cancer Imaging Archive},
   title = {Data From NSCLC-Radiomics (version 4) [Data set]},
   year = {2014}
}

@article{Molino2019,
   author = {Piero Molino and Yaroslav Dudin and Sai Sumanth Miryala},
   month = {9},
   title = {Ludwig: a type-based declarative deep learning toolbox},
   url = {http://arxiv.org/abs/1909.07930},
   year = {2019}
}

@article{Cardoso2022,
   author = {M. Jorge Cardoso and Wenqi Li and Richard Brown and Nic Ma and Eric Kerfoot and Yiheng Wang and Benjamin Murrey and Andriy Myronenko and Can Zhao and Dong Yang and Vishwesh Nath and Yufan He and Ziyue Xu and Ali Hatamizadeh and Andriy Myronenko and Wentao Zhu and Yun Liu and Mingxin Zheng and Yucheng Tang and Isaac Yang and Michael Zephyr and Behrooz Hashemian and Sachidanand Alle and Mohammad Zalbagi Darestani and Charlie Budd and Marc Modat and Tom Vercauteren and Guotai Wang and Yiwen Li and Yipeng Hu and Yunguan Fu and Benjamin Gorman and Hans Johnson and Brad Genereaux and Barbaros S. Erdal and Vikash Gupta and Andres Diaz-Pinto and Andre Dourson and Lena Maier-Hein and Paul F. Jaeger and Michael Baumgartner and Jayashree Kalpathy-Cramer and Mona Flores and Justin Kirby and Lee A. D. Cooper and Holger R. Roth and Daguang Xu and David Bericat and Ralf Floca and S. Kevin Zhou and Haris Shuaib and Keyvan Farahani and Klaus H. Maier-Hein and Stephen Aylward and Prerna Dogra and Sebastien Ourselin and Andrew Feng},
   month = {11},
   title = {MONAI: An open-source framework for deep learning in healthcare},
   url = {http://arxiv.org/abs/2211.02701},
   year = {2022}
}


\end{document}